\documentclass[sn-mathphys, pdflatex]{sn-jnl}% arxiv

\usepackage{amsmath}
\usepackage{amsfonts}
\usepackage{gensymb}
\usepackage{multirow}
\usepackage{subfig}
\usepackage{tabularx}
\usepackage{booktabs}
\usepackage{comment}
\usepackage{xcolor}
\usepackage{xspace}

\newcommand{\trans}{T}
\newcommand{\ie}{\textit{i}.\textit{e}.}
\newcommand{\taun}[1]{\ensuremath{\|\tau_{#1}\|_2}}
\newcommand{\phin}[1]{\ensuremath{\|\phi_{#1}\|_2}}
\newcommand{\disp}{\ensuremath{D}}
\newcommand{\seg}{\ensuremath{S}}
\newcommand{\res}{\ensuremath{\Delta j_{t+1}}}
\newcommand{\src}{\ensuremath{i_t}}
\newcommand{\tgt}{\ensuremath{j_{t+1}}}
\newcolumntype{Y}{>{\centering\arraybackslash}X}

\newcommand{\name}{TAToo\xspace}
\usepackage{indentfirst}

\jyear{2023}%

\theoremstyle{thmstyleone}%
\theoremstyle{thmstyletwo}%

\theoremstyle{thmstylethree}%

\begin{document}

\title[Video-based Anatomy and Tool Tracking]{\name: Vision-based Joint Tracking of Anatomy and Tool for Skull-base Surgery}

%%=============================================================%%
%% Prefix	-> \pfx{Dr}
%% GivenName	-> \fnm{Joergen W.}
%% Particle	-> \spfx{van der} -> surname prefix
%% FamilyName	-> \sur{Ploeg}
%% Suffix	-> \sfx{IV}
%% NatureName	-> \tanm{Poet Laureate} -> Title after name
%% Degrees	-> \dgr{MSc, PhD}
%% \author*[1,2]{\pfx{Dr} \fnm{Joergen W.} \spfx{van der} \sur{Ploeg} \sfx{IV} \tanm{Poet Laureate} 
%%                 \dgr{MSc, PhD}}\email{iauthor@gmail.com}
%%=============================================================%%

\author[1]{\fnm{Zhaoshuo} \sur{Li}} 
% \email{zli122@jhu.edu}

\author[1]{\fnm{Hongchao} \sur{Shu}} 
\author[1,2]{\fnm{Ruixing} \sur{Liang}} 
\author[1]{\fnm{Anna} \sur{Goodridge}}
\author[1]{\fnm{Manish} \sur{Sahu}}
\author[2]{\fnm{Francis X.} \sur{Creighton}}
\author[1]{\fnm{Russell H.} \sur{Taylor}}
\author[1]{\fnm{Mathias} \sur{Unberath}}

\affil[1]{\orgname{Johns Hopkins University}, \orgaddress{\city{Baltimore}, \state{MD}, \country{U.S}}}
\affil[2]{\orgname{Johns Hopkins Medicine}, \orgaddress{\city{Baltimore}, \state{MD}, \country{U.S}}}

%%================================%%
%% Sample for structured abstract %%
%%================================%%

\abstract{
\textbf{Purpose:} Tracking the 3D motion of the surgical tool \textit{and} the patient anatomy is a fundamental requirement for computer-assisted skull-base surgery. 
The estimated motion can be used both for intra-operative guidance and for downstream skill analysis. 
Recovering such motion solely from surgical videos is desirable, as it is compliant with current clinical workflows and instrumentation.
\newline

\textbf{Methods:} We present Tracker of Anatomy and Tool (\name).
\name jointly tracks the rigid 3D motion of the patient skull and surgical drill from stereo microscopic videos. 
\name estimates motion via an iterative optimization process in an end-to-end differentiable form.
For robust tracking performance, \name adopts a probabilistic formulation and enforces geometric constraints on the object level.
\newline

\textbf{Results:} We validate \name on both simulation data, where ground truth motion is available, as well as on anthropomorphic phantom data, where optical tracking provides a strong baseline. 
We report sub-millimeter and millimeter inter-frame tracking accuracy for skull and drill, respectively, with rotation errors below 1$\degree$. 
We further illustrate how \name may be used in a surgical navigation setting.
\newline

\textbf{Conclusion:} We present \name, which simultaneously tracks the surgical tool and the patient anatomy in skull-base surgery. 
\name directly predicts the motion from surgical videos, without the need of any markers. 
Our results show that the performance of \name compares favorably to competing approaches. 
Future work will include fine-tuning of our depth network to reach a 1\,mm clinical accuracy goal desired for surgical applications in the skull base.}

\keywords{
Image-based navigation, 3D motion tracking, Computer vision, Deep learning, Computer-assisted interventions{\let\thefootnote\relax\footnote{{Code is available at: \url{https://github.com/mli0603/TAToo}}}}
}

%%\pacs[JEL Classification]{D8, H51}

%%\pacs[MSC Classification]{35A01, 65L10, 65L12, 65L20, 65L70}

\maketitle

\section{Introduction}\label{sec1}
Many otologic and neurosurgical procedures require that surgeons use a surgical drill to remove bone in the lateral skull-base to gain access to delicate structures therein.  This process requires both high precision and a strong visual understanding of the anatomy in order to avoid damage to critical anatomies.  If the \textit{rigid} 3D motion of both surgical drill and patient skull can be tracked relative to each other and provided to the surgeons, we can potentially improve the operation safety through this assistive system~\cite{mezger2013navigation}. The recovered 3D motion can also be used to assess surgical skill \cite{azari2019modeling} in post-operative analysis. 

Despite the widespread use of external tracking systems such as optical trackers, video-based solutions are desirable since they integrate seamlessly into existing surgical workflows. Furthermore, for the purposes of skill analysis, they would enable a retrospective analysis of videos that were acquired without specialized external tracking instrumentation. 
However, tracking multiple objects in surgical video streams is challenging due to the need for the identification of individual objects and maintaining cross-frame correspondences for motion estimation \cite{braspenning2004true}. 
Prior video-based systems~\cite{liu2022sage,speers2018fast,long2021dssr, wang2022neural} focus on tracking the patient anatomy with respect to the camera but disregard other objects of interest, such as surgical tools. 
This restricts their applicability to skull-base surgery, where the rigid motion of both the patient skull and surgical drill are of interest. 
Other work \cite{lee2017multi,gsaxner2021inside} assumes no modification of the patient anatomy and requires precise 3D shapes, which is inapplicable to skull-base surgery.

\begin{figure}[b]
    \centering
    \includegraphics[width=\linewidth]{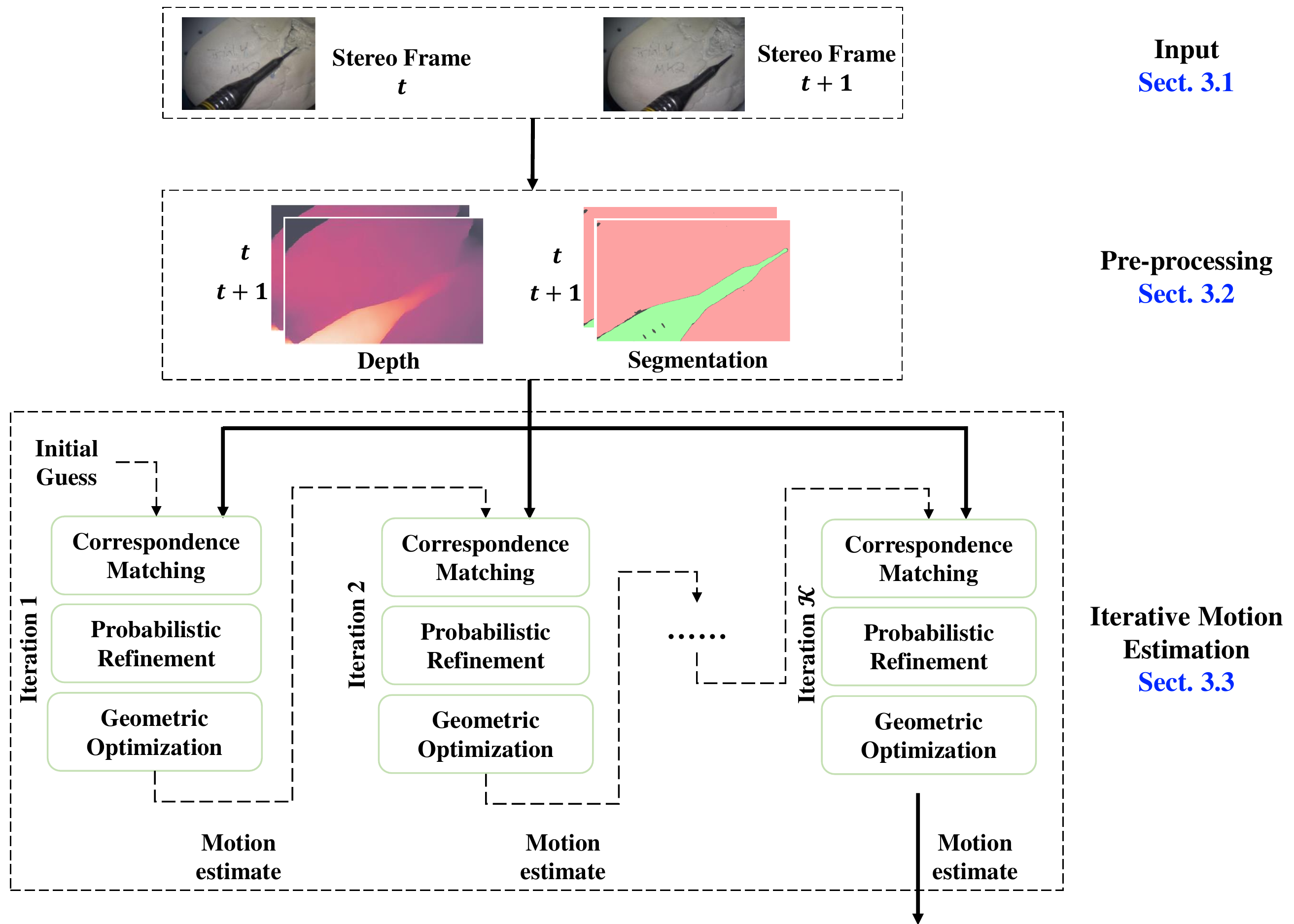}
    \caption{Overview of \name estimating the 3D motion for patient skull and surgical drill. \name takes two stereo video frames across time as input (\autoref{ssec:input}), pre-processes the frames  (\autoref{ssec:preprocess}) and uses an iterative motion estimation process to regress the motion (\autoref{ssec:motion_estimation}).}
    \label{fig:overview}
\end{figure}

We introduce \name (\autoref{fig:overview}), which simultaneously tracks the rigid 3D motion of the patient skull and surgical drill relative to the microscope, without prior 3D information of the surgical scene. 
Given a stereo video stream as input, \name first uses off-the-shelf networks to estimate the stereo depth~\cite{li2021revisiting, li2021temporally, tankovich2021hitnet} and segmentation map~\cite{shvets2018automatic} as a pre-processing step. 
\name then iteratively updates the motion estimate of both patient skull and surgical drill. 
At each iteration, \name matches the correspondences, refines the correspondences in a probabilistic formulation, and lastly regresses consistent object motion based on geometric optimization. 
The whole method is end-to-end differentiable. 

For evaluation, we specifically consider a skull-base surgical procedure named mastoidectomy, where the temporal bone is drilled. 
As no public dataset is available for our intended application, we collect a set of data for developing \name using both simulation and anthropomorphic phantom data emulating the surgical setup (see \autoref{fig:experiment_setup}). 
We benchmark \name against other motion estimation techniques, including keypoint- and ICP-based algorithms. 
\name's performance compares favorably to competing approaches, and we further find it to be more robust. 
We show that \name achieves sub-millimeter and millimeter tracking accuracy for patient skull and surgical drill, respectively, with a rotation errors below $1\degree$. 
We lastly illustrate how \name may be used in a surgical navigation setting.
Our contributions can be summarized as follows:
\begin{itemize}
    \item We present a novel framework, \name, that tracks the 3D motion of patient skull and surgical drill jointly from stereo microscopic videos.
    \item We demonstrate that the recovered 3D motion from \name can be used in downstream applications, such as a surgical navigation setting.
\end{itemize}

\section{Related Work For 3D Motion Estimation}
Motion estimation in general requires matching correspondences temporally and regressing 3D motions based on the matches. \textbf{Keypoint-based approaches} detect keypoints from images \cite{rublee2011orb}, match keypoints across frames, and then estimate the motion using Procrustes-type registration \cite{kabsch1976solution}. In surgical scenes, detecting keypoints can be challenging due to large homogeneous regions. Even if keypoints can be found, the sparsity of the detected points can lead to poor spatial configurations that are not adequate for motion estimation. 
\textbf{ICP-based approaches} \cite{besl1992method,park2017colored} instead iteratively find correspondences based on the most recent motion estimate and a set of distance criteria. While the correspondence is often dense and object-level rigidity is enforced, ICP is sensitive to outliers \cite{zhang2021fast}, which occur often during stereo depth estimation or segmentation estimation. Our approach \name builds upon \cite{besl1992method,teed2021raft,teed2021tangent} to use all-pixel correspondences with a probabilistic formulation for improved performance. \name further enforces object-level constraints based on semantic image segmentation for effective motion tracking of multiple objects. 

\section{Approach}
\subsection{Input}
\label{ssec:input}
We denote the left and right stereo images at a given time as \textit{one stereo frame}. Given stereo frames at time $t$ and $t+1$ from a microscopic video stream, \name recovers the 3D motion of the patient skull and the surgical drill from $t$ to $t+1$ with respect to the left stereo camera. We use $H$ and $W$ to denote image height and width, subscripts $p$ for \textit{p}atient skull, and $d$ for surgical \textit{d}rill. We use $\trans \in \mathbb{SE}(3)$ for rigid 3D motion. The output of \name is $T_p$ and $T_d$, where for convenience of notation we have omitted the temporal dependence.

\subsection{Pre-processing}
\label{ssec:preprocess}
We first use off-the-shelf networks \cite{tankovich2021hitnet,shvets2018automatic} to estimate the depth and segmentation information and the associated prediction probabilities. We denote the depth map as $\disp$ and the probability of the estimate as $\sigma(\disp)$. The segmentation map denoted as $\seg$, groups pixels into different objects. Each pixel $i\in HW$ belongs to either patient skull ($\seg^i=p$) or surgical drill ($\seg^i=d$). The probabilities of the segmentation assignment are denoted as $\sigma(\seg)$. The depth and segmentation information is estimated for both frames at $t$ and $t+1$.

\subsection{Iterative Motion Estimation}
\label{ssec:motion_estimation}
Given an initial guess of the motion of the skull and drill, \name iterates between correspondence matching (\autoref{sssec:motion_feature}), probabilistic refinement (\autoref{sssec:recurrent_network}), and geometric optimization (\autoref{sssec:geometric_opt}). In our work, the initial guess of the motion is set to be zero (\ie, the identity transformation). 

\subsubsection{Correspondence Matching}
\label{sssec:motion_feature}
Given the most recent 3D motion estimate, we first compute the resulting correspondences across frames.

For each \textit{source} pixel $\src \in H_t W_t$ from frame $t$, we compute its \textit{target} location $\tgt$ in frame $t+1$ given the most recent motion estimate. The correspondence pair is thus formed as $\src, \tgt$. To compute $\tgt$, we use either the patient skull motion $\trans_p$ or drill motion $\trans_d$ according to the segmentation prediction $\seg_t^i$:
\begin{equation}
    \tgt = \pi \big( \trans \pi^{-1}(\src) \big), \,\, \trans = \begin{cases}
    \trans_p , & \text{if} \,\, \seg^i_t = p,\\
    \trans_d , & \text{if} \,\, \seg^i_t = d,
  \end{cases} 
\end{equation}
where $\pi$ and $\pi^{-1}$ are perspective and inverse perspective projections for conversion between pixel and Cartesian world coordinates. 

\begin{figure}[t]
    \centering
    \includegraphics[width=\linewidth]{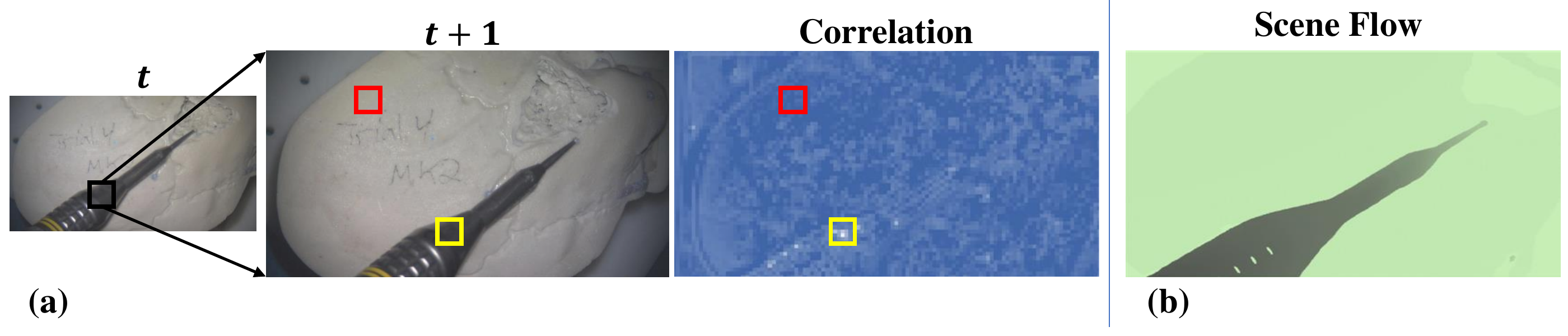}
    \caption{Qualitative visualization of the motion features. (a) We visualize an $1 \times 1 \times H_{t+1} \times W_{t+1}$ slice of the 4D correlation volume. This slice of correlation corresponds to the correlation between a source pixel in frame $t$ (black) and all pixels in frame $t+1$.  \textit{Brighter} color indicates \textit{larger} correlation, \ie, the yellow target pixel has a larger correlation with the source pixel than the red target pixel. (b) Scene flow between correspondence pairs.}
    \label{fig:motion_features}
\end{figure}

\subsubsection{Probabilistic Refinement}
\label{sssec:recurrent_network}
As the correspondence pairs computed from the most recent motion estimate contain outliers, we perform a refinement on these correspondence pairs and estimate the confidence of the refined results.
Following lietorch~\cite{teed2021tangent}, we use a deep learning network to learn how to perform such refinement.
An illustration of the network design is in Appendix A.

Given a correspondence pair $\src,\tgt$, we first evaluate the fitness of the match. 
Our network extracts features from the \textit{left} images of both stereo frames and builds a 4D correlation volume \cite{teed2021raft} of $H_t \times W_t \times H_{t+1} \times W_{t+1}$, where the dot-product correlation between all features in frame $t$ and all features in frame $t+1$ are evaluated. 
A larger value in the correlation volume indicates a more probable match. 
We retrieve the \textit{correlation} between $\src$ and $\tgt$ from the 4D correlation volume in \autoref{ssec:preprocess}. We also compute the resulting \textit{scene flow} as $\hat{f}^i=\tgt-\src$. 
Both the correlation values and scene flow are used for refinement prediction. 
A qualitative visualization is shown in \autoref{fig:motion_features}.
 
Using the correlation and scene flow as input, the network then updates the correspondence pair by predicting a \textit{residual update} $\res$ to the target location $\tgt$ while fixing the source location $\src$:
\begin{equation}
\hat{j}_{t+1} = \tgt + \res \, . 
\end{equation}
In order to evaluate how confident the network is about the update, we also output a probability of such residual update via a sigmoid layer. The refinement probability is denoted as $\sigma(\res)$. 

Given the updated correspondence pair $\src$ and $\hat{j}_{t+1}$, we compute the \textit{joint probability} $\sigma(\src,\hat{j}_{t+1})$ of the matching. 
Intuitively, this joint probability indicates the reliability of the current correspondence pair. 
We decompose the joint probability $\sigma(\src,\tgt)$ into two terms: 1) the confidence of the information we know about the source point $\sigma(\src)$, 2) and confidence of the information we know about the target point $\sigma(\hat{j}_{t+1} \vert \src)$:
\begin{equation}
    \sigma(\src,\hat{j}_{t+1}) = \underbrace{\sigma(\seg^i_t) \sigma(\disp^i_t)}_{\sigma(\src) \text{:\,estimate probability}} 
    \,\,\,\,\,
    \underbrace{\sigma(\res) \sigma(\seg^j_{t+1}) \sigma(\disp^j_{t+1}) \,,}_{\sigma(\hat{j}_{t+1}  \vert \src) \text{:\,correspondence probability}}
    \label{eqn:joint_prob}
\end{equation}
where $\sigma(\seg^i_t),\sigma(\disp^i_t)$ are the confidence of depth and segmentation at $\src$ from frame $t$, and $\sigma(\seg^j_{t+1}),\sigma(\disp^j_{t+1})$ are the confidence of depth and segmentation at target $\hat{j}_{t+1}$ from frame $t+1$. With a slight abuse of notation, the correspondence probability $\sigma(\hat{j}_{t+1} \vert \src)$ is written as a conditional probability because the target locations $\hat{j}_{t+1}$ are computed from $\src$. 
Our network uses a GRU design with convolution layers following prior work on recurrent optimization \cite{teed2021raft}.

\subsubsection{Geometric Optimization}
\label{sssec:geometric_opt}
Given the matched points and the associated confidence, we regress the motion of both the skull and tool with geometric constraints. 
We employ Gauss-Newton optimization steps over the $\mathbb{SE}(3)$ space, following its recent success in
single object non-rigid tracking \cite{bozic2020neural}, structure from motion \cite{lindenberger2021pixel}, and SLAM \cite{teed2021tangent}. We estimate motion by minimizing the perspective projection error between the target location $\hat{j}_{t+1}$ and the transformed pixel locations from $\src$, weighted by the joint probabilities in \autoref{eqn:joint_prob}:
\begin{equation}
    E(\trans) = \sum_{\src \in H_t W_t} \sigma(\src,\hat{j}_{t+1}) \cdot \|\hat{j}_{t+1} - \pi \big( \trans \pi^{-1}(\src) \big) \|_2, \,\, \trans = \begin{cases}
    \trans_p , & \text{if} \,\, \seg^i_t = p,\\
    \trans_d , & \text{if} \,\, \seg^i_t = d,
    \end{cases}
    \label{eqn:energy}
\end{equation} 
where $\trans_p, \trans_d$ are optimized. The intuition behind \autoref{eqn:energy} is that given the perspective projection relationship, as well as the network predicted correspondence and probability, it finds the motion $\trans_p$ and $\trans_d$ that best explain the predicted correspondences $\src,\hat{j}_{t+1}$.

\subsection{Supervision} 
We adopt a trained off-the-shelf depth network~\cite{tankovich2021hitnet} for estimating depth information.
We fine-tune a segmentation network \cite{shvets2018automatic} on our dataset. 

We then train the network in the motion estimation process (\autoref{sssec:recurrent_network}) and impose a loss on the geodesic distance~\cite{teed2021tangent} between ground truth and predicted motion for each object on the Lie manifold of $\mathbb{SE}(3)$:
\begin{align}
    \ell_\text{geo} &= \taun{p} + \phin{p} + \taun{d} + \phin{d}, \\ 
    \tau, \phi &= \log(T^\text{GT} \trans^{-1}),
    \label{eqn:error}
\end{align}
where $\trans^\text{GT}$ is the ground truth motion, $\tau$ is the translation vector, and $\phi$ is the Rodrigues' rotation vector. 

Given the ground truth motion, we also derive the match locations $\tgt^\text{GT}$ and the scene flow $f^{i,GT}$ for additional supervision, and thus, impose losses as:
\begin{align}
     \ell_\text{match} = \frac{1}{H_t W_t }\sum_{\src \in H_t W_t} \vert \tgt^\text{GT} - \tgt \vert \,, \\
    \ell_\text{flow} = \frac{1}{H_t W_t }\sum_{\src \in H_t W_t} \vert f^{i,GT}-f^i \vert \,.
\end{align}
We note that $\ell_\text{match}$ is used to supervise recurrent network incremental estimates $\res$, and $\ell_\text{flow}$ is used to supervise the motion estimate from the geometric optimization. The refinement probability $\sigma(\res)$ is implicitly learned without any supervision.

The above loss is computed for estimates at each iteration. The summed loss is weighted differently at each optimization iteration:
\begin{equation}
    \ell = \sum_{k\in[1,\mathcal{K}]} 0.3^{\mathcal{K}-k} (w_\text{geo}\ell_\text{geo,k} + w_\text{match} \ell_\text{match,k} + w_\text{flow} \ell_\text{flow,k}) \,,
\end{equation}
where the final iteration is weighted most. We set $w_\text{geo}=10.0$, $w_\text{flow}=0.1$ and $w_\text{match}=0.1$ to balance loss magnitudes.

\begin{figure}[tb]
    \centering
    \includegraphics[width=\linewidth]{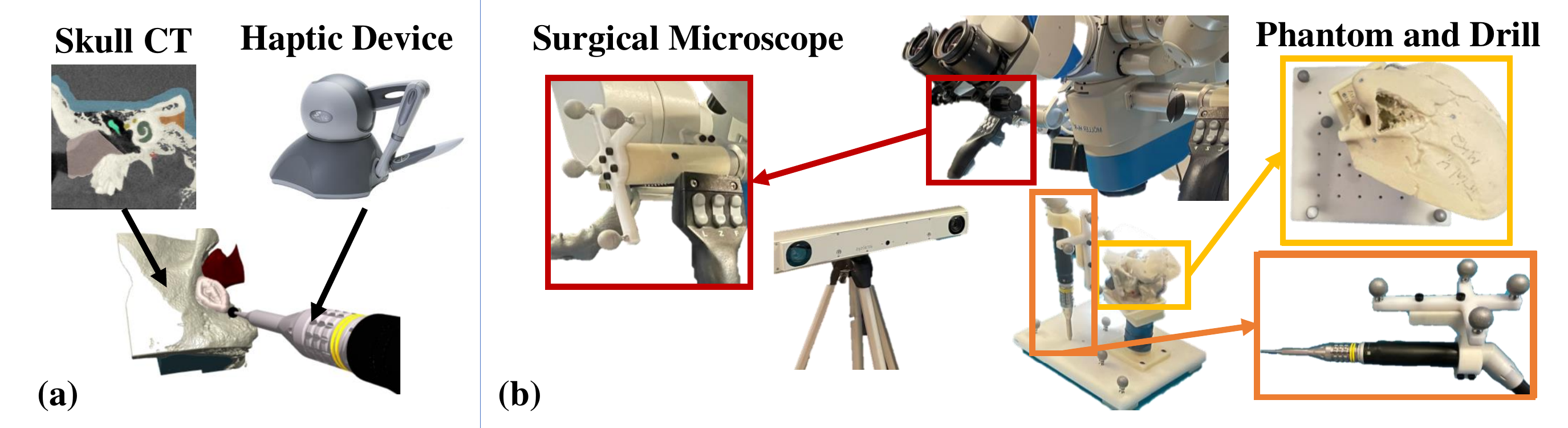}
    \caption{Data collection setup: (a) simulation environment, (b) surgical phantom with optical tracking as the baseline.}
    \label{fig:experiment_setup}
\end{figure}

\section{Experimental Setup}
\subsection{Data}
\textbf{Simulation} We use a drilling simulator \cite{munawar2021virtual, munawar2023fully} to generate synthetic data of three different CT scans \cite{ding2021automated, ding2022automated, ding2023self} drilled by surgical residents as shown in \autoref{fig:experiment_setup}(a). For each CT scan, 1500 instances of data are recorded (4500 total). The simulation data contains ground truth depth, segmentation, and motion. The image resolution is $640\times480$. We use two sequences for training/validation and one sequence for testing. 
\newline

\textbf{Phantom} We additionally collect four video sequences of surgical phantom data as shown in \autoref{fig:experiment_setup}(b). We use the Atracsys fusionTrack optical tracker\footnote{\url{https://www.atracsys-measurement.com/products/fusiontrack-500/}} and mount retro-reflective tracking markers on the surgical microscope, phantom, and drill to acquire individual poses, which are used to compute inter-frame motion. There were a total of 13915 instances of data. The image resolution is $960\times540$. We use three sequences for training/validation and one sequence for testing. 

\subsection{Training and Evaluation Setup}
We train \name on synthetic data and then fine-tune on surgical phantom data. The initial motion estimate is set to be the identity transformation. We use pre-trained depth network~\cite{tankovich2021hitnet} since we do not have ground truth depth in the phantom data. We manually annotate 100 frames to fine-tune the segmentation network \cite{shvets2018automatic} for the surgical phantom data. We set the number of iterations to $\mathcal{K}=3$. We use random cropping and color augmentations during the training process. We further sub-sample and also reverse the video frames to augment motion data. We use a 80-20 train-validation split ratio. 

For evaluation, we benchmark different motion regression algorithms taking the depth and segmentation estimates as a given input. We report the mean L2 norm of translation and rotation error vectors (\autoref{eqn:error}) and threshold metrics of 1\,mm and $1\degree$ over the entire video sequence. We compare our motion estimation technique against a keypoint-based approach using ORB features and brute force matching \cite{rublee2011orb}, and the colored ICP algorithm \cite{park2017colored} implemented by Open3D\footnote{\url{http://www.open3d.org/}}.

\section{Results and Discussion}

\subsection{Tracking Accuracy}

\indent
\autoref{tab:benchmark} summarizes the results from both synthetic and surgical phantom data. 
In both cases, our method outperforms competing approaches by a large margin. 

\begin{figure}[t]
    \centering
    \includegraphics[width=\linewidth]{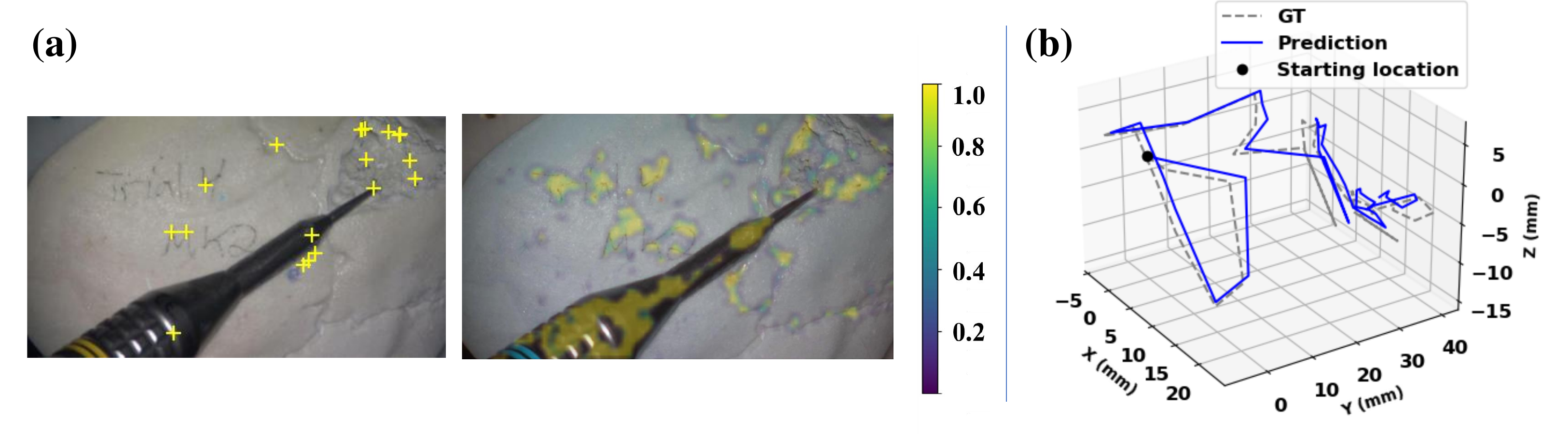}
    \caption{(a) Spatial distribution of keypoints (yellow crosses, left) and correspondence probabilities $\sigma(\src,\hat{j}_{t+1})$ of our method (colormap, right) both overlaid on frame $t$. Our probable correspondences are dense and distributed, which is better conditioned for motion estimation. (b) The plot of surgical drill trajectory in patient coordinate as used in a surgical navigation system.}
    \label{fig:qual_vis}
\end{figure}

The keypoint-based approach performs worst due to the sparsity and poor spatial configuration of matches, especially for the surgical drill. 
When there are fewer than 3 keypoints detected, the keypoint-based approach fails to recover the motion, resulting in high failure rates on both datasets.
Even for the surgical phantom, many keypoints are clustered in local patches, which is undesirable for motion estimation.
In contrast, ICP and \name both use dense correspondences to avoid such failure cases.
We visualize the matches found by our method in \autoref{fig:qual_vis}(a), where the correspondences are evenly distributed across the objects.

\begin{figure}[t]
    \centering
    \includegraphics[width=\linewidth]{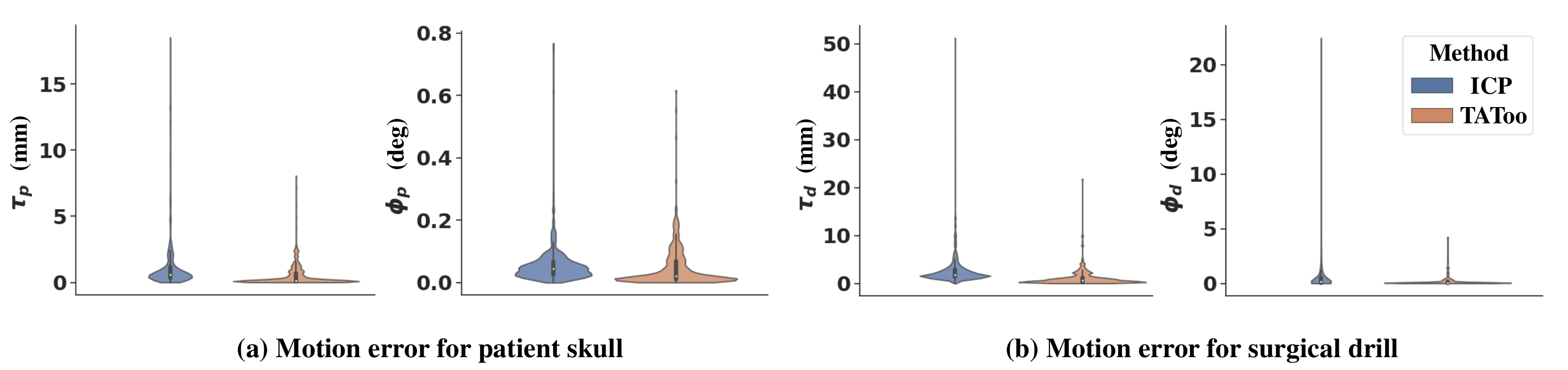}
    \caption{Violin plots of motion errors comparing ICP and \name. \name contains much fewer outliers in motion prediction than ICP due to the probabilistic formulation to reject outliers.}
    \label{fig:error_plot}
\end{figure}

ICP is also inferior to \name due to insufficient outlier rejection during correspondence search, since rejection is based solely on distance and color \cite{zhang2021fast}.
In contrast, \name uses a fully probabilistic formulation for the matched correspondences, considering confidences in depth, segmentation, and matches to regress the motion.
Thus, \name can mitigate the effect of outliers and estimate motion more robustly. 
We show the violin plots of motion errors in \autoref{fig:error_plot} to demonstrate that \name indeed contains fewer extreme outliers in prediction due to the improved robustness.

On both data, our approach has better tracking performance for the patient skull than the surgical drill. This is attributed to the larger inter-frame motion of the surgical drill. Indeed, the average motion is 0.1\,mm and $0.03\degree$ for the skull, but 1.1\,mm and $0.38\degree$ for the drill. 

While our approach compares favorably to other techniques, we note that our performance deteriorates on surgical phantom data compared to synthetic data, especially for surgical drill tracking. We attribute this deterioration, at least partially, to the sim-to-real transfer challenge in the stereo depth estimation network, as we do not have ground truth data to fine-tune the model.
We ablate the impact of depth and segmentation quality on the motion tracking accuracy and present the results in Appendix B.
Qualitative results of the depth and segmentation estimates are shown in Appendix C.
It is our future work to investigate techniques to alleviate the sim-to-real transfer issue. 

\setlength\tabcolsep{0.6em}
\begin{table}[t]
\centering
\caption{Benchmark result on synthetic and surgical phantom data. For all metrics, lower is better. $\| \tau \|_2$: translation error. $\| \phi \|_2$: rotation error. Failure rate: percentage of the video where motion cannot be recovered.}
\label{tab:benchmark}
\begin{tabularx}{\linewidth}{c | cc | cc | c}
\toprule
 &
  \multicolumn{5}{c}{\textbf{Synthetic Data}} \\
 &
  \taun{p} (mm) &
  \phin{p} (\degree) &
  \taun{d} (mm) &
  \phin{d} (\degree) &
  Failure Rate \\ \hline
Keypoint & 
    29.9 $\pm$ 34.2 &
    2.4 $\pm$ 2.8 &
    10.8 $\pm$ 7.8 &
    34.0 $\pm$ 38.5 &
    19\% \\ 
 ICP &
    1.5 $\pm$ 2.5 &
    0.1 $\pm$ 0.1 &
    2.9 $\pm$ 3.9 &
    4.2 $\pm$ 25.0 &
    0\% \\ 
 \textbf{\name (ours)} &
    0.5 $\pm$ 0.9 &
    0.1 $\pm$ 0.1 &
    1.1 $\pm$ 1.8 &
    0.2 $\pm$ 0.4 &
    0\% \\\toprule
 &
  \multicolumn{5}{c}{\textbf{Surgical Phantom Data}} \\
 &
  \taun{p} (mm) &
  \phin{p} (\degree) &
  \taun{d} (mm) &
  \phin{d} (\degree) &
  Failure Rate \\ \hline
  Keypoint & 
    7.5 $\pm$ 4.8 &
    0.8 $\pm$ 0.5 &
    10.3 $\pm$ 2.03 &
    16.5 $\pm$ 36.5 &
    30\%  \\ 
ICP &
    0.7 $\pm$ 0.7 &
    0.1 $\pm$ 0.1 &
    9.7 $\pm$ 4.5 &
    2.0 $\pm$ 13.9 &
    0\%
   \\ 
\textbf{\name (ours)} &
    0.2 $\pm$ 0.2  &
    0.1 $\pm$ 0.1 &
    4.8 $\pm$ 3.5 &
    0.7 $\pm$ 0.8 &
    0\% \\
   \bottomrule
\end{tabularx}
\end{table}

\subsection{Downstream Application - Surgical Navigation}
We apply our approach to surgical navigation, where inter-frame motion predictions are chained for absolute poses. We report the average drill-to-skull transformation error over the entire video sequence on the validation set of synthetic data. The average error is 3.6\,mm and $0.3\degree$, demonstrating the applicability of the \name video-based tracking paradigm in such a setting. 
A qualitative visualization is shown in \autoref{fig:qual_vis}(b) and additional visualization can be found in the video supplementary material. 
While promising, accumulating relative pose for navigation inevitably introduces drift. 
Additional mechanisms, such as pose graph bundle adjustment, are required to meet the often referenced clinical requirement of $<$1\,mm accuracy \cite{schneider2021evolution}.

\section{Conclusion}
We present a stereo video-based 3D motion estimation approach that simultaneously tracks the patient skull and the surgical drill.
Our proposed iterative optimization approach combines learning-based correspondence matching with geometric optimization and probabilistic formulation. 
Experiments on simulation and phantom data demonstrate that our approach outperforms competing motion estimation methods. 

While we show promising results, our evaluation is limited to simulation and phantom data. We plan to further improve our methods, collect \textit{in-vivo} dataset and expand our analysis.
Moreover, while \name outperforms other image-based tracking algorithms, the accuracy of \name does not currently meet our 1 mm clinical accuracy goal, which we attribute to the lack of fine-tuning on the depth and segmentation network.
Future work includes collecting frame-wise ground truth data for supervising our depth and segmentation network using scalable data collection setup such as digital twins~\cite{shu2023twin}.

% \backmatter

% \bmhead{Supplementary information}

% If your article has accompanying supplementary file/s please state so here. 

% Authors reporting data from electrophoretic gels and blots should supply the full unprocessed scans for key as part of their Supplementary information. This may be requested by the editorial team/s if it is missing.

% Please refer to Journal-level guidance for any specific requirements.
\bmhead{Acknowledgments} This work was supported in part by Johns Hopkins University internal funds, and in part by NIDCD K08 Grant DC019708.

% Acknowledgments are not compulsory. Where included they should be brief. Grant or contribution numbers may be acknowledged.

% Please refer to Journal-level guidance for any specific requirements.

\section*{Declarations}
\bmhead{Conflict of interest} Dr. Russell H. Taylor and Johns Hopkins University may be entitled to royalty payments related to technology discussed in this paper, and Dr. Taylor has received or may receive some portion of these royalties. Also, Dr. Taylor is a paid consultant to and owns equity in Galen Robotics, Inc. These arrangements have been reviewed and approved by Johns Hopkins University in accordance with its conflict of interest policy.

\bibliography{reference}% common bib file
%% if required, the content of .bbl file can be included here once bbl is generated
%%\input sn-article.bbl

%% Default %%
%%\input sn-sample-bib.tex%

\end{document}